\documentclass{article}

\usepackage{arxiv}

\usepackage[utf8]{inputenc} 
\usepackage[T1]{fontenc}    
\usepackage{hyperref}       
\usepackage{url}            
\usepackage{booktabs}       
\usepackage{amsfonts}       
\usepackage{nicefrac}       
\usepackage{microtype}      
\usepackage{natbib}
\usepackage{doi}

\usepackage[T1]{fontenc}
\usepackage{times}%
\usepackage{graphicx}
\usepackage{tabulary}

\title{Automatic generation of a large dictionary with concreteness/abstractness ratings based on a small human dictionary}

\author{ Vladimir Ivanov \\
	Faculty of Computer Science and Software Engineering \\
	Innopolis University \\
	1, Universitetskaya st., Innopolis \\
	Republic of Tatarstan, 420500 \\
	\And
	Valery Solovyev \\
	Linguistic research and education center \\
	Research laboratory `Intellectual technologies of text management' \\
	Kazan Federal University \\
	2, Tatarstan Street, Kazan \\
	Republic of Tatarstan, 420021 \\
}

\renewcommand{\shorttitle}{\textit{arXiv} Template}

\begin{document}
\maketitle

\begin{abstract}
Concrete/abstract words are used in a growing number of psychological and neurophysiological research. For a few languages, large dictionaries have been created manually. This is a very time-consuming and costly process. To generate large high-quality dictionaries of concrete/abstract words automatically one needs extrapolating the expert assessments obtained on smaller samples. The research question that arises is how small such samples should be to do a good enough extrapolation. In this paper, we present a method for automatic ranking concreteness of words and propose an approach to significantly decrease amount of expert assessment. The method has been evaluated on a large test set for English. The quality of the constructed dictionaries is comparable to the expert ones. The correlation between predicted and expert ratings is higher comparing to the state-of-the-art methods.
\end{abstract}




\section{Introduction}
Large dictionaries with assessments of the semantic and psycholinguistic properties of words are an important support for experimental and quantitative research in the cognitive sciences, psychology, and linguistics \cite{solovyev2021ca}. Dictionaries containing tens of thousands of words provide a selection of the most appropriate words for research. Unfortunately, the creation of such dictionaries is very expensive and time consuming as it needs experts. 

Expert assessments of the various semantic properties of words are used in a wide range of studies from theoretical psychology to word processing. The most popular are assessments of the degree of valence, arousal, dominance, age-of-acquisition, and concreteness. 

In this article, we will deal with the concreteness/abstractness property. First, let us look into definitions of concreteness and abstractness. The main approach to defining these concepts presented in \cite{spreen1966parameters}.
Concrete concepts are those that are perceived by the senses. Examples of concrete words are \textit{`cat', `chair', `mountain'}. Abstract concepts are not perceived by the senses. For example, \textit{`responsibility', `relationships', `misunderstanding'}. Similar interpretations are found in many works. For instance, the following definition is given in the paper \cite{schmid2012english}
\textit{``abstract nouns are those nouns whose denotata are not part of the concrete physical world and cannot be seen or touched''}.

For these purposes, dictionaries (databases) are created by collecting word scores from respondents. The psycholinguistic database assembled by \cite{coltheart1981mrc} became the first large publicly available digital resource of this kind for English. Recently, crowdsourcing platforms such as Amazon’s Mechanical Turk have been actively used to enable  evaluating large number of words. An important example of a dictionary with ratings of concreteness/abstractness (C/A ratings) of words\footnote{We will denote such dictionaries in abbreviated form as `C/A dictionaries'} for English is the dictionary created by \cite{brysbaert2014concreteness}, which contains estimates of almost 40 thousand words and phrases.
Many studies use the 40-thousand dataset for English. The original paper \cite{brysbaert2014concreteness} is cited almost 1000 times in Google Scholar. In other works (e.g. in \cite{hollis2017extrapolating}) the need for a large-scale data collection to support research within psychology is highlighted.

Except English language, a dictionary containing tens of thousands of words with expert assessments of concreteness/abstractness exists only for the Dutch language \cite{brysbaert2014norms}. In this regard, the problem of automatic generation of C/A dictionaries is open for many natural languages. Moreover, Hollis et al. \cite{hollis2017extrapolating} doubt that human ratings provide a good gold standard and discuss a possibility of replacing manual ratings with unambiguous, empirically accessible measures. An obvious requirement for automatic dictionaries is high correlation with human dictionaries. In \cite{mandera2015useful}, authors discuss possible errors in computer dictionaries and note that the higher the correlation between an automatic dictionary and a human one, the less likely significant errors will appear in the automatic dictionary.

The main idea behind the extrapolation of expert assessments to words without C/A score, is to use the semantics learned from a large corpus and to obtain new estimates based on the semantic similarity of words in a vector space built from a text corpus. Thus, to create an automatic dictionary in any language, it is necessary to have a large corpus of texts (in that language). On this basis, pre-trained vector representations of words (word embeddings) can be built.
Major ideas to leverage pre-trained word embeddings in building a C/A dictionary can be reduced to the following two approaches.
\begin{enumerate}
  \item Using an estimator (such as regressor or classifier, e.g. a neural network). The estimator is trained on a portion of the expert assessments and then tested on the rest of the dictionary.
  \item Using a small sample of ``reference'' abstract and concrete words (called `a core' or `a seed'). A C/A rating for a given word is calculated by a comparison of similarity to the both parts of the core.
\end{enumerate}

Most researchers take the first approach. It is very convenient in view of the development of the theory and technology of machine learning, especially taking into account the progress in deep learning. An obvious disadvantage of this approach is that it needs a lot of data to train classifiers, therefore initial dictionaries with expert judgements should be large. Only a few languages have dictionaries with thousands of words. Thus, for the vast majority of languages, this approach is not available.

In contrast, the second approach does not need a large train set. The core can be quite small, so that manual creation is quite realistic for any language. However, there is currently neither a general theory for the creation of such core words, nor empirical studies related to selection of the size of a seed. The importance of seed selection, its size and affect on the resulting dictionary have been studied by \cite{mandera2015useful}: ``If an optimal set of seed words would increase the accuracy of the extrapolation methods, it would be good to know this''. In our work, we try to give an answer to this question.

Within the framework of the two mentioned approaches, C/A ratings can be generated for arbitrarily large dictionaries, as long as word embeddings are available for a given language. For example, \cite{turney2011literal} built a dictionary of 114,501 words; \cite{koper2016automatically} created even bigger one (350,000 words). However, a common problem remains to assess the quality of the dictionaries obtained in this way. Almost all algorithms are tested on small collections of several thousand words, and it is unclear how the algorithms will perform on large collections of words. Most of the existing resources are either created for the most frequent words or have other limitations on input words. For example, The Toronto Word Pool \cite{friendly1982toronto}, used in \cite{charbonnier2019predicting}, contains only fairly frequent common words of the English language, containing a maximum of 2 syllables or 8 letters. A dictionary with C/A ratings created in \cite{brysbaert2014concreteness} also contains the most common, frequent English words.
As more frequent words have more contexts in corpus, building C/A dictionaries them is easier, than for less frequent words.
In contrast, estimating C/A ratings for low-frequency words (on average) will be worse because language models can capture fewer contexts of such words. In \cite{solovyev2020generation}, this assumption was confirmed on the material of Russian language.

In present article, we analyze a large number of semantic cores (or seeds) on the basis of a large dictionary proposed in \cite{brysbaert2014concreteness} which contains human assessments for over 39 thousand words. We study general rules for selecting the size of the semantic C/A core. The C/A ratings for the seed words can be derived by a survey of respondents-experts. The main finding of the paper is that the amount of data collected in such a survey (that is enough to generate a high-quality C/A dictionary) may be tens of words (not hundreds or thousands). In addition, for the first time we apply contextualized word embeddings to extrapolate C/A ratings, and show that contextualized word embeddings (ELMo and BERT) provide better source for semantic core selection then `classical' word embeddings (e.g. fastText).
Finding a small semantic core for this problem is directly related to the fundamental problem of reducing the size of the training set for tuning neural networks.

\section{Related Works}

\label{sec:related}

Most of the work on the extrapolation of human ratings was carried out on the material of English using two main digital resources \cite{brysbaert2014concreteness,coltheart1981mrc}. Among the first works, one can mention \cite{theijssen2011difficulty} with the correlation coefficient (Spearman's rank coefficient, $r_s$) between the dictionary and the human ratings being 0.64. The Table \ref{sota-table} below summarizes results of studies carried out within the classification-based approach. In some works, instead of building a dictionary with ratings, a binary classifier is built to distinguish between concrete and abstract words. In this case, instead of Spearman's correlation coefficient ($r_s$), the accuracy of binary classification is calculated. 

\begin{table*}
\centering
\begin{tabulary}{\linewidth}{LLLLLLLL}
\hline \textbf{Paper}  & \textbf{Corpus} & \textbf{Semantic space} & \textbf{Method} & \textbf{Train size} & \textbf{Split}& \textbf{Performance} \\ \hline

\cite{mandera2015useful} &  \cite{brysbaert2014concreteness} & LSA 
& kNN, RF & 37,058 & 25\% /75\% & .796 ($r_s$) \\ 

\hline
\cite{feng2011simulating} &  \cite{coltheart1981mrc} & LSA & SVR & 3,521 & 67\% /33\% & .802 ($r_s$) \\ 
\hline
\cite{tsvetkov2013cross} &  \cite{coltheart1981mrc} & vector space & LR  & 2,450 & 98\% /2\% & .76 (acc.) \\ 
\hline 
\cite{hollis2017extrapolating} &  \cite{brysbaert2014concreteness} & skip-gram & SVR & 37,058 & 50\% /50\% & .829 ($r_s$)\\ 
\hline 

\cite{rabinovich2018learning} &  \cite{brysbaert2014concreteness} & GloVe & NB, RNN, kNN & 2,580  & 81\% /19\% & .740 ($\rho$) \\ 
\hline 
\cite{ljubevsic2018predicting} &  \cite{brysbaert2014concreteness} & fastText & SVR & 22,797 & 67\% /33\% & .887 ($r_s$) \\ 
\hline
\cite{charbonnier2019predicting} & \cite{brysbaert2014concreteness} & fastText & SVM & 32,783 & 90\% 
/10\% & .900 ($r_s$) \\ 

\hline
\cite{charbonnierpredicting2020} & \cite{charbonnierpredicting2020}* & fastText & SVR & 4,182 &  80\% /20\% & .861 ($r_s$) \\ 
\hline 

\end{tabulary}
\caption{Summary of studies carried out within the classification-based approach (in the chronological order).  Asterisk (*) denotes German language.}
\label{sota-table} 
\end{table*}

Here are important notes about the papers from Table \ref{sota-table}. The works \cite{mandera2015useful,rabinovich2018learning} present a comparison of different methods. Authors of  
\cite{rabinovich2018learning} trained a neural network on sentences, and not on separate words (800,000 sentences with 2,580 words annotated as abstract and concrete were used). 
Aiming at exploiting both embeddings and textual data, authors utilized a bidirectional recurrent neural network (RNN) with one layer of forward and backward LSTM cells. Each cell has width of 128, and is wrapped by a dropout wrapper with keep probability 0.85. The output of the LSTM cells is passed to the attention layer which reduces it to the size of 100. GloVe embeddings with 300 dimensions were used as word representations. Given a set of sentences containing a test concept, its final abstractness score was computed by applying the averaging based on the Naive Bayes classifier.

In \cite{charbonnier2019predicting}, several extrapolations are given, including those using additional linguistic information for training classifiers, such as a list of suffixes typical for abstract words. For comparison with the results of other studies, the table shows the results of the algorithm without using this additional information.

Note that in almost all works, except \cite{mandera2015useful}, the test set is much smaller than the training set (in \cite{hollis2017extrapolating} it is equal to the training set in size). The possibilities of applying the algorithms described in these papers to the rating of large dictionaries are not clear, because they vastly depend on the generalizing ability of a proposed model. Although, \cite{mandera2015useful} use a large test set (almost 30 thousand words), the training set is also very large; it is more than 9 thousand words. Accordingly, the possibilities of applying the algorithm in those languages in which there are only small dictionaries with human estimates seem limited. The state-of-the-art result is $r_s=0.9$. In 2014 \cite{brysbaert2014concreteness} estimated the correlation of two human ratings of works \cite{coltheart1981mrc} and \cite{brysbaert2014concreteness}. The value of Spearman's $r_s$ was $0.92$. Later, in 2017 \cite{hollis2017extrapolating} have suggested that this estimate can be treated as the upper limit for the quality estimates of other automatic ratings.

Several papers have used a small core of concrete and abstract words. The results are shown in Table \ref{sota-table2}. In all the works, a greedy algorithm was used to build the core, and the core size, following the work \cite{turney2011literal}, was a priori set equal to 40 words.
Table \ref{sota-table2} shows the size of dictionary in which the core was searched as the fraction of a training set. Within the framework of this approach, sufficiently high correlation coefficients were obtained. However, it should be noted that, for example, \cite{charbonnierpredicting2020} use the size of the set in which the core is sought is more than three thousand words, while the test set was slightly more than 800 words. Thus, this approach retains the same drawbacks as the one with classifiers. It should be noted that in all works that use the core, the influence of either the core size or influence of the training/test set size were not studied.

\begin{table*}[t]
\centering
\begin{tabulary}{\linewidth}{LLLlllLL}
\hline \textbf{Paper}  & \textbf{Corpus} & \textbf{Semantic space}  & \textbf{Train size} & \textbf{Split} & \textbf{Performance} \\ \hline
\cite{turney2011literal} &	\cite{coltheart1981mrc} &	LSA	  &	4,295 & 50\% /50\% &	.810 ($r_s$);  .847 (acc.) \\
\hline
\cite{koper2016automatically}&	\cite{koper2016automatically}* &	word2vec 	 &	5,237 & 90\% /10\% &	.825 ($r_s$) \\ 
\hline
\cite{charbonnierpredicting2020}  &	\cite{charbonnierpredicting2020}* &	fastText  &	4,182 & 80\% /20\% &	.849 ($r_s$) \\

\hline
\end{tabulary}
\caption{\label{sota-table2} Papers that use core-based approach, i.e. use a small core with concrete and abstract words. Asterisk (*) denotes German language. All works apply `greedy forward search' when searching for a better core.}
\end{table*}

Let us note several noteworthy results of the recent works, in addition to those given in the table \ref{sota-table2}.
The work done by \cite{snefjella2019historical} stands apart; authors extrapolate ratings in a diachronic manner, building a dictionary for a 200-year interval.
In works \cite{thompson2018automatic,ljubevsic2018predicting}, extrapolation is carried out not within one language, but between languages. \cite{thompson2018automatic} use a multilingual skip-gram model. In this case, the interpolation method is trained on the full set of available data of one language. \cite{thompson2018automatic} transfer assessments to 77 languages, but data for all the languages are not provided. When transferring estimates from English to Dutch, the correlation coefficient with expert estimates in Dutch from \cite{brysbaert2014norms} turned out to be 0.76. 

When starting the algorithm from a small core (as in works \cite{snefjella2019historical,turney2011literal}), the question arises about the choice of words in the core. In \cite{snefjella2019historical}, the core of a fixed size included the most frequent and at the same time extremely concrete and abstract, according to expert estimates, words. In \cite{turney2011literal} there is also a core of a predetermined size of forty words. 
It is formed iteratively, starting from an empty set and sequentially adding words that are closest in semantic space with abstract or concrete words from the training. It is shown that when the core is expanded to 100 words, the correlation coefficient on the test set drops.


\cite{feng2011simulating} have shown that that words with higher concreteness ratings were more likely to be categorized as artifacts, foods, animals, people, substances, plants, or body parts. Less concrete words were more likely to be categorized as related to cognition, action, shapes, communication, relations, states, events, time, or motives. In addition, in the same work, authors made the following statement: ``... we hypothesized that more concrete words would be less ambiguous (i.e., less polysemous). This was not the case.''

Automatic ratings are constructed for several other languages: for Chinese \cite{wang2018method}, for Persian \cite{dadras2017codac}, for Russian \cite{vv2019dictionary,solovyev2020automated}.

\section{Data and Methods}

\subsection{Datasets and performance measures}
In our experiments we use a dictionary with expert C/A ratings, the BRM dictionary for English from \cite{brysbaert2014concreteness}. The BRM dictionary contains 39,954 words and phrases (bigrams). It is widely used in studies of concreteness/abstractness estimation. The dictionary has C/A ratings ranging from 1 (most abstract) to 5 (most concrete) for 37,058 unigrams and 2,896 bigrams. Average C/A rating is 3.04.

Second type of resource used in experiments is the frequency dictionaries. For English we use the dictionary of 1/3 million most frequent words available online\footnote{https://norvig.com/ngrams/}, for more information on the dataset, see Chapter 14 in \cite{SegaranHammerbacher2009}. The frequency dictionary overlaps with the BRM dictionary and we combine them, because having access to word frequencies is important for the proposed method. 

Finally, we employ three pre-trained word embeddings to calculate similarity between words: 
fastText \cite{bojanowski2017enriching}, ELMo \cite{Peters:2018} and BERT \cite{devlin2018bert} (for BERT-based experiments we use bert-base-uncased model and aggregate outputs of last layers of the encoder). Vector representations for words are derived via the Flair Framework \cite{akbik2018coling}. FastText representations of words use subword n-grams, which is useful for C/A prediction, because concreteness/abstractness of a word can depend not only on the context of the word, but also on subword information. 

Usage of fastText-based method enables its application to many languages (for which fastText representations exist), while pre-trained contextualized word embeddings (based on BERT encoder) exist for a  smaller number of languages. 
To assess the performance of fastText embeddings we compare them to ELMo and to a BERT-based representations of words. 
The corresponding datasets and models are freely available. We made this choice due to the availability of the word embeddings in many languages.


To evaluate the performance of C/A ratings estimation we use three common metrics: Spearman's rank correlation ($r_s$), Pearson correlation coefficient ($\rho$) and accuracy of binary classification ($Acc$). The former two metrics can be evaluated on the results of the regression task (predicting C/A rating of a word), while the accuracy (rate of correct predictions) is measured on the binary classification task. 

\subsection{Methods for estimating C/A ratings}
We follow a general semantic-based approach used in the above works and do not deal with syntactic information, POS-tags, etc.

\subsubsection{Semantic core method}

Here we briefly describe the method based on a semantic core.
The key aspect of the method is selection of a semantic core, that is formed by two sets of the abstract and concrete words (with known C/A ratings). The second step of the method is just calculating similarities between word embeddings in a semantic space. 
Therefore, a rigorous analysis of semantic core selection and optimization of the core is the main contribution of the paper, while using the semantic space is secondary.

First, we define a semantic core, which consists of two sets of words: $seed_A$ with $Z$ abstract words and $seed_C$ with $Z$ concrete words\footnote{Further, we refer to the two sets as `seeds'. Also, we use terms `concrete seed' and `abstract seed' to refer to the corresponding set.}. 
$$\texttt{semantic core} = seed_A \cup seed_C$$

Next, given a word $w$, the method retrieves an embedding for the word $w$ and calculates cosine similarity between the word $w$ and words from both seeds. The result of this step is average similarity between the word $w$ and each seed. We assume the following: the closer the word $w$ to the concrete seed ($seed_C$); and farther it is from the abstract seed ($seed_A$), the higher the concreteness rank of the word $w$ should be. The corresponding concreteness score of word $w$ can be calculated with the following formulas:
$$Rating(\mathbf{w}) =  \frac{sim(\mathbf{w},seed_C)}{sim(\mathbf{w},seed_A)}$$
where 
$$sim(\mathbf{w},seed_C) = \frac{1}{|seed_C|} \sum_{\mathbf{c} \in seed_C} \frac{\mathbf{w} \cdot \mathbf{c}}{\Vert \mathbf{w} \Vert \Vert \mathbf{c} \Vert}  $$
The cosine similarity is traditionally used in this research domain. Other measures of similarity lead to like results.
The score of the word $w$ depends on the selection of the semantic core. Therefore, selection and optimization of the seeds is the important step of the method that worth studying. 

\subsubsection{Selection and optimization of the semantic core}
Selection of the semantic core is possible from a larger dictionary. In this case, we fix several parameters such as size of the larger dictionary ($X$) and two subsets of larger dictionary (each of size $Y$). Then seeds are sampled from the subsets ($X > Y > Z$). This approach resembles real-world situation, when usually one experts are able to collect C/A ratings of $X$ words, but want to minimize the effort, especially, if $Z$ words would be enough to build a high-quality dictionary. 

\begin{enumerate}
\item Take in it the sub-dictionary of the most frequent words of size $X$ (Base dictionary).
\item We select two subsets (each of size $Y$) of the most concrete and the most abstract words.
\item Then, we randomly choose two seeds (each of size $Z$)\footnote{In particular, $Z$ words from the concrete subset comprise a concrete seed, while other randomly selected $Z$ words from the abstract subset comprise an abstract seed. After this step we have a semantic core with $2Z$ words.}. 
\item Using this core, we calculate $Rating(\textbf{w})$ for each word from the Base dictionary.
\item We calculate Spearman's correlation ($r_s$) between the test and the predicted ratings.
\item Search for better $X, Y, Z$ and build a complete dictionary for the derived core. (Optimization step)
\item Optimal values $X, Y, Z$ are used for testing on the entire test dictionary (calculate $r_s$ between the test dictionary and predictions for all words test dictionary).
\end{enumerate}

The Step 6 above mentions searching for a better (in terms of $r_s$) core. In this optimization step, the goal is to find a better semantic core. We apply a brute-force strategy that randomly picks and evaluates seeds. 
\begin{enumerate}
\item We start with initial values for $X$ (500 words)
\item Set up $Y$=50 words.
\item We iterate over all core sizes $Z$ greater than ten words and less than or equal to $Y$.
\item For each $Z$, we iterate over all cores of size $Z$ (if the number of all combinations is too big, then we stop the search after 100 random cores) and keep the best core.
\item Increase $Y$ with an increment of 50 up to $X / 3$. Repeat steps 3-4.
\item Increase $X$ with an increment of 500 up to 2,500. Repeat steps 2-5.
\end{enumerate}

Selection of the initial and maximum values for $X$ in the optimization strategy is justified as follows. Apart from a few large expert C/A dictionaries, the majority of them have between 1,000 and 2,700 words. For example, a dictionary for German includes 1,698 words \cite{wippich2013bildhaftigkeit}. The database \cite{lahl2009using} contains 2,654 words. The database \cite{kanske2010leipzig} includes 1,000 nouns. For English language the database \cite{friendly1982toronto} contains 1,080 words, a dictionary from \cite{paivio1968concreteness} contains 925 words. This indicates real limitations on the size of the expert C/A dictionaries. Based on these data, we made a decision to study methods of automatic dictionary construction, provided that 500--2,500 words with expert assessments are available.

\section{Experiments}
\begin{figure*}
    \centering
    \includegraphics[width=14cm]{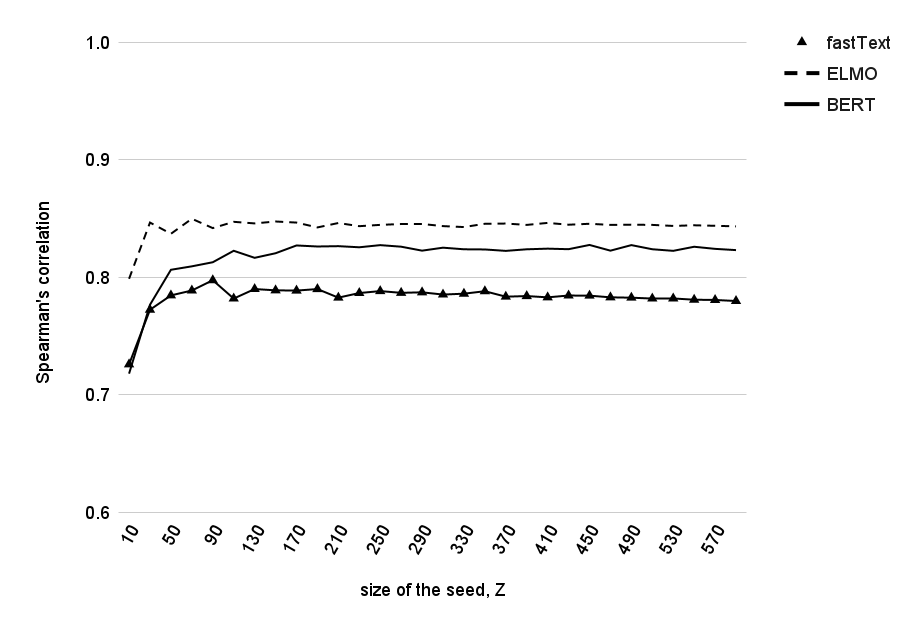}
    \caption{Comparison of results obtained for different types of embeddings and for different values of $Z$ with fixed values of $X=2500$ and $Y=600$.}
    \label{fig:corr}
\end{figure*}

The experimental results are summarized in Tables \ref{core-table-1}--\ref{small-core-table-3} and presented in Fig. \ref{fig:corr}. We will start considering the results obtained with the fastText-based model (Tables \ref{core-table-1}, \ref{small-core-table-1}).

\begin{table}
\centering
\begin{tabulary}{\linewidth}{llllL}
\hline 
$X$ & $Y$ & $Z$ & $r_s$ & Difference from the best\\ 
 \hline
500 & 50 & 30 & 0.710 & 10\% \\
1000 & 300 & 130 & 0.763 & 4\% \\
1500 & 200 & 50 & 0.775 & 3\% \\
2000 & 400 & 70 & 0.790 & 1\% \\
2500 & 600 & 90 & \textbf{0.797} & 0 \\

\hline
\end{tabulary}
\caption{\label{core-table-1} Comparison of the best results and different values of $X$ for fastText.}
\end{table}

\begin{table}
\centering
\begin{tabulary}{\linewidth}{llllL}
\hline 
$X$ & $Y$ & $Z$ & $r_s$ & Difference from the best\\ 
 \hline
 500 &	150 &	110 &	0.800 & 6.5\% \\
 1000 &	300 &	30 &	0.825 & 3.5\% \\
 1500 &	350 &	30 &	0.842 & 1.5\% \\
2000 &	350 &	30 &	0.848 & 0.8\% \\
2500 &	200 &	30 &	\textbf{0.855} & 0 \\

\hline
\end{tabulary}
\caption{\label{core-table-2} Comparison of the best results and different values of $X$ for ELMo.}
\end{table}

\begin{table}
\centering
\begin{tabulary}{\linewidth}{llllL}
\hline 
$X$ & $Y$ & $Z$ & $r_s$ & Difference from the best\\ 
 \hline
 500 &	150 &	110 &	0.788 & 5.3\% \\
 1000 &	100 &	90 &	0.829 & 0.4\%\\
 1500 &	150 &	50 &	0.825 & 0.8\%\\
 2000 &	600 &	210 &	0.829 & 0.4\%\\
2500 &	550 &	270 &	\textbf{0.832} & 0 \\

\hline
\end{tabulary}
\caption{\label{core-table-3} Comparison of the best results and different values of $X$ for BERT.}
\end{table}
From the Table \ref{core-table-1} one can see the following:
\begin{itemize}
    \item The more vocabulary available, the better
    \item The difference between levels of 2000 words and 2500 words is less less than 1\%
    \item To obtain a dictionary of an acceptable level of quality with the least effort, it is enough to create an expert dictionary of 1000 words ($X=1000$)
    \item The optimal value of $Y$ is on average 20\% of $X$. The optimal value of $Z$ is on average 25-30\% of $Y$.
\end{itemize}

\begin{table}
\centering
\begin{tabulary}{\linewidth}{lll}
\hline 
 $Y$ & $Z$ &  $r_S$ \\
 \hline
300 & 130 & \textbf{0.763}\\
100 & 50 & 0.760 \\
150 & 70 & 0.760 \\
250 & 70 & 0.759 \\
100 & 70 & 0.759 \\

\hline
\end{tabulary}
\caption{\label{small-core-table-1} Five best values of correlation at $X = 1000$ for fastText.}
\end{table}

\begin{table}
\centering
\begin{tabulary}{\linewidth}{lll}
\hline 
 $Y$ & $Z$ &  $r_S$ \\
 \hline
300 &	30 &	\textbf{0.825} \\ 
150 &	50 &	0.824 \\
100 &	70 &	0.824 \\
100 &	90 &	0.823 \\
300 &	210 &	0.822 \\

\hline
\end{tabulary}
\caption{\label{small-core-table-2} Five best values of correlation at $X = 1000$ for ELMo.}
\end{table}

\begin{table}
\centering
\begin{tabulary}{\linewidth}{lll}
\hline 
 $Y$ & $Z$ &  $r_S$ \\
 \hline
100 &	90 &	\textbf{0.829} \\
100 &	70 &	0.828 \\
250 &	170 &	0.827 \\
250 &	190 &	0.826 \\
200 &	70 &	0.823 \\
\hline
\end{tabulary}
\caption{\label{small-core-table-3} Five best values of correlation at $X = 1000$ for BERT.}
\end{table}


In this table, the average is $Y = 180, Z = 78$. Recommended values for use in practice are $Y = 100, Z = 50$. These setting gives a correlation coefficient that differs from the best by no more than 5\%. 
For reasonable values of the value of the expert vocabulary (e.g. up to 2500 words, a core with parameters $X = 2500, Y = 600, Z = 90$ was found, giving the best result 0.797. Figure \ref{fig:corr} illustrates the graph of correlation coefficient obtained with different values of $Z$. There is a maximum (at $Z=90$) It is clear that increasing of $Z$ does not improve the quality of the C/A dictionary. Similar trend is derived for other combinations of $X$ and $Y$ values.

The recommendation is to limit the creation of a 1000-word dictionary, which will allow obtaining the correlation coefficient only 4\% worse than the best. When using the exhaustive search algorithm proposed in the article and limited resources, it is recommended to limit the search procedure by a seeds of size $Z=50$, with $Y = 100$, which gives a result that differs from the best by 5\% (Table \ref{small-core-table-1}).Similar patterns were found for ELMo (Tables \ref{core-table-2}, \ref{small-core-table-2}) and for BERT (Tables \ref{core-table-3}, \ref{small-core-table-3}). 

Comparing the results for these three models, we see that ELMo provides the absolute best result (0.855). At the same time, if we initially limit ourselves to a dictionary of 1000 words, then the ELMo and BERT results are very close, and the BERT result (0.829) which is slightly better. The fastText results are noticeably worse.
Figure \ref{fig:corr} shows relationship between core size ($Z$) and $r_S$ for all three models. The three plots are qualitatively the same. The ELMo-based model reaches a local maximum already for a core size 30, and the further growth of the core size does not improve results.
We provide words from the typical seeds with reasonable size in Table \ref{cores}.

\begin{table}[!ht]
\centering
\begin{tabulary}{\linewidth}{L|L}
\hline Concrete seed & Abstract seed \\ \hline
`shoe', 
`clock', 
`bracelet', 
`computer', 
`bird', 
`bed', 
`bean', 
`pantyhose', 
`neck', 
`oven'	&
`desire', `moment', `reliability', `opportunity', `choice', `concept', `value', `peace', `sensitivity', `democracy' \\
\hline
\end{tabulary}
\caption{\label{cores} Example of seeds obtained from fastText semantic space with following parameters: $X=2500, Y=100, Z=10$; $r_s = 0.764$ }
\end{table}





\section{Conclusion}
\label{sect:concl}

A number of studies have addressed the problem of extrapolating human C/A scores to words that do not have such a rating. 

It is fundamentally important to be able to extend human assessments from a small initial set of words to a much larger one. Almost all the previous work, as well as our estimates, were obtained by dividing the initial data set into parts (e.g. 80-90\% for training set and 20-10\% for test set). Meanwhile, only for two languages (English and Dutch) large human C/A dictionaries for tens of thousands of words have been built. The problem of building a large C/A dictionary for other languages based on a small set of words with human ratings is an urgent problem. 

Even smaller sets of initial data are used in the approach associated with the choice of the core of concrete and abstract words and the subsequent calculation of the distance from the evaluated word to the words of the core. 
Despite its simplicity, the proposed method is able to calculate C/A scores that strongly correlate with human ratings. 

However, to date, no systematic study of issues related to the choice of the core has been undertaken. Using the brute-force method, we evaluated a large number of core selection situations, varying in several parameters, including the size of the source data and the size of the core.

The research is carried out for the main word embeddings models: ELMo, BERT and fastText. FastText has the advantage of having pre-trained vectors for 170 languages, which makes it useful for building machine dictionaries for languages other than the most common. However, if there is a pre-trained ELMo or BERT for the language, then our research shows that it is better to use them.

Comparing with other works in which the core-based approach was used, it can be noted that we obtain a result slightly higher than the best previously published one: the correlation coefficient is 0.855 versus 0.849 in \cite{charbonnierpredicting2020}. We have the required core size of 30, which is also better than the best previous result with a core size of 40 \cite{turney2011literal}. Earlier, in articles with a core-based approach, various variants of the semantic space were used: LSA, word2veec, fasttext.

We found a core of size 30 that (in combination with the ELMo semantic space and the limitation of 2,500 initially specified words) gave the result 0.855 ($r_S$). At the same time, it was shown that initial dictionary can be even smaller (1,000 words)
that leads to $r_S=$0.825, which is only a small (3.5\%) deterioration of the result in comparison with the best found. 

Comparing our results with previous works, one should pay attention to the following. Within the framework of the core-based approach, the best result was obtained in \cite{charbonnierpredicting2020} for the German language (0.849). At the same time, a set of words of a larger size was used for training (3,300 words). For the English language, the best result (0.810) presented in \cite{turney2011literal} when using a dictionary with 2,150 words. In the framework of the classification-based approach, the best result is 0.9 ($r_S$), but it requires a large training set of almost 30,000 words, which is currently available only for two languages. But for these two languages (English and German), a problem of constructing a large set with C/A ratings is already solved and the machine approximation to solve this problem is not relevant.

Dependencies of the result on the parameters under consideration have been established, which generally clarifies the structure of the semantic space of words from the standpoint of concreteness/abstractness. With a fixed size of the original set of words with human ratings, the core size has some optimal value, and as the core size increases, the results deteriorate. Earlier, a similar dependence was announced, but not published in works using other semantic models and methods for constructing the core. Thus, it seems that a certain universal pattern has been established, independent of the specific technique used.

In general, the conclusion from this study is a recommendation for those languages where C/A dictionaries have not yet been built, to limit themselves to building a small human dictionary and then apply  computational method to automatically extrapolate estimates to a large dictionary. It is an opportunity to quickly and inexpensively get a high-quality C/A dictionary. 
The study demonstrates (and, for the first time, validates) a possibility to derive large high-quality language resources (C/A ratings) using a small training set. Usage of the simple (or even trivial) methods is definitely an advantage of the approach that enables its application to many low-resource languages. 

Our future plans include conducting similar studies for other semantic characteristics of words as well as estimation of C/A ratings for other languages.



\section*{Acknowledgements}
This paper has been supported by the Kazan Federal University Strategic Academic Leadership Program.







\bibliographystyle{acl_natbib}
\bibliography{lke2021}

\end{document}